%% file: bare_conf.tex
\documentclass[10pt, conference, compsocconf]{IEEEtran}
\usepackage{ifpdf}

\usepackage{makecell}
\usepackage{multirow} 
\usepackage{tabularx} 
\usepackage{array}    

\usepackage{cite}

\ifCLASSINFOpdf
  \usepackage[pdftex]{graphicx}
\else
  \usepackage[dvips]{graphicx}
  
\fi
\usepackage[cmex10]{amsmath}
\usepackage{amssymb}
\usepackage{algorithmic}

\usepackage{array}

\usepackage{mdwmath}
\usepackage{mdwtab}

\usepackage{eqparbox}

\usepackage{subcaption}

\usepackage{caption}
\usepackage{fixltx2e}

\usepackage{stfloats}

\usepackage{url}

\begin{document}

%
\title{Dense Monocular Motion Segmentation Using Optical Flow and Pseudo Depth Map: A Zero-Shot Approach}


\author{\IEEEauthorblockN{Yuxiang Huang}
\IEEEauthorblockA{Systems Design Engineering\\
University of Waterloo\\
Waterloo, Canada\\
yuxiang.huang@uwaterloo.ca}
\and
\IEEEauthorblockN{Yuhao Chen}
\IEEEauthorblockA{Systems Design Engineering\\
University of Waterloo\\
Waterloo, Canada\\
yuhao.chen1@uwaterloo.ca}
\and
\IEEEauthorblockN{John Zelek}
\IEEEauthorblockA{Systems Design Engineering\\
University of Waterloo\\
Waterloo, Canada\\
jzelek@uwaterloo.ca}
}


%


\maketitle

\input{0_abstract}    
\input{1_introduction}
\input{2_related_work}

\input{4_methodology}
\input{5_experiments}
\input{6_conclusions}

%
\IEEEpeerreviewmaketitle

\clearpage
\bibliography{MoSeg_Ref}
\bibliographystyle{IEEEtran}

%



\end{document}

%% file: 0_abstract.tex
\begin{abstract}
Motion segmentation from a single moving camera presents a significant challenge in the field of computer vision. This challenge is compounded by the unknown camera movements and the lack of depth information of the scene. While deep learning has shown impressive capabilities in addressing these issues, supervised models require extensive training on massive annotated datasets, and unsupervised models also require training on large volumes of unannotated data, presenting significant barriers for both. In contrast, traditional methods based on optical flow do not require training data, however, they often fail to capture object-level information, leading to over-segmentation or under-segmentation. In addition, they also struggle in complex scenes with substantial depth variations and non-rigid motion, due to the overreliance of optical flow. To overcome these challenges, we propose an innovative hybrid approach that leverages the advantages of both deep learning methods and traditional optical flow based methods to perform dense motion segmentation without requiring any training. Our method initiates by automatically generating object proposals for each frame using foundation models. These proposals are then clustered into distinct motion groups using both optical flow and relative depth maps as motion cues. The integration of depth maps derived from state-of-the-art monocular depth estimation models significantly enhances the motion cues provided by optical flow, particularly in handling motion parallax issues. Our method is evaluated on the DAVIS-Moving and YTVOS-Moving datasets, and the results demonstrate that our method outperforms the best unsupervised method and closely matches with the state-of-the-art supervised methods.

\end{abstract}


%% file: 1_introduction.tex
\section{Introduction}
\label{sec:intro}

Dense motion segmentation aims to partition a video frame into separate areas characterized by consistent motion. The ability to identify and segment moving objects with a moving camera is crucial for various downstream applications, including autonomous navigation, robotics, simultaneous localization and mapping (SLAM), and holistic scene understanding. In dynamic environments, the video camera moves at an unknown speed relative to its environment, posing significant challenges for motion segmentation methods, such as motion parallax and motion degeneracy \cite{Hartley2004}. 

Existing techniques for dense monocular motion segmentation can be divided into two main categories: traditional methods which heavily rely on the brightness constancy constraint (usually in the form of optical flow) for the motion cue, and deep learning methods that can learn multiple motion and appearance cues from the video sequence through training. However, both types of methods have inherent limitations. Deep learning based approaches can produce impressive results on complex scenes where there are multiple moving objects, different types of motions and significant depth variations, but the state-of-the-art deep learning methods require end-to-end training with a significant amount of supervision, which is computation intensive and does not generalize well to different scenes \cite{mohamed_monocular_2021, vertens_smsnet_2017, ramzy_rst-modnet_2019, dave_towards_2019, yang_learning_2021, neoral_monocular_nodate}. Conversely, most traditional methods rely heavily on the optical flow or the brightness constancy constraint, which significantly limits their performance in complex scenes with significant depth variations or non-rigid motions \cite{sekkati_variational_2007, cremers_detection_2009, papazoglou_v_2013, fragkiadaki_video_2012, keuper_motion_2015}. Moreover, traditional methods are not able to accurately delineate moving objects as a whole \cite{keuper_motion_2015, leibe_its_2016}, especially when there are multiple moving components of the same object, due to their inability to learn the high-level appearance information of the objects. 

\begin{figure*} [!t] 
    \centering
    \begin{subfigure}{\textwidth}
        \includegraphics[width=\linewidth]{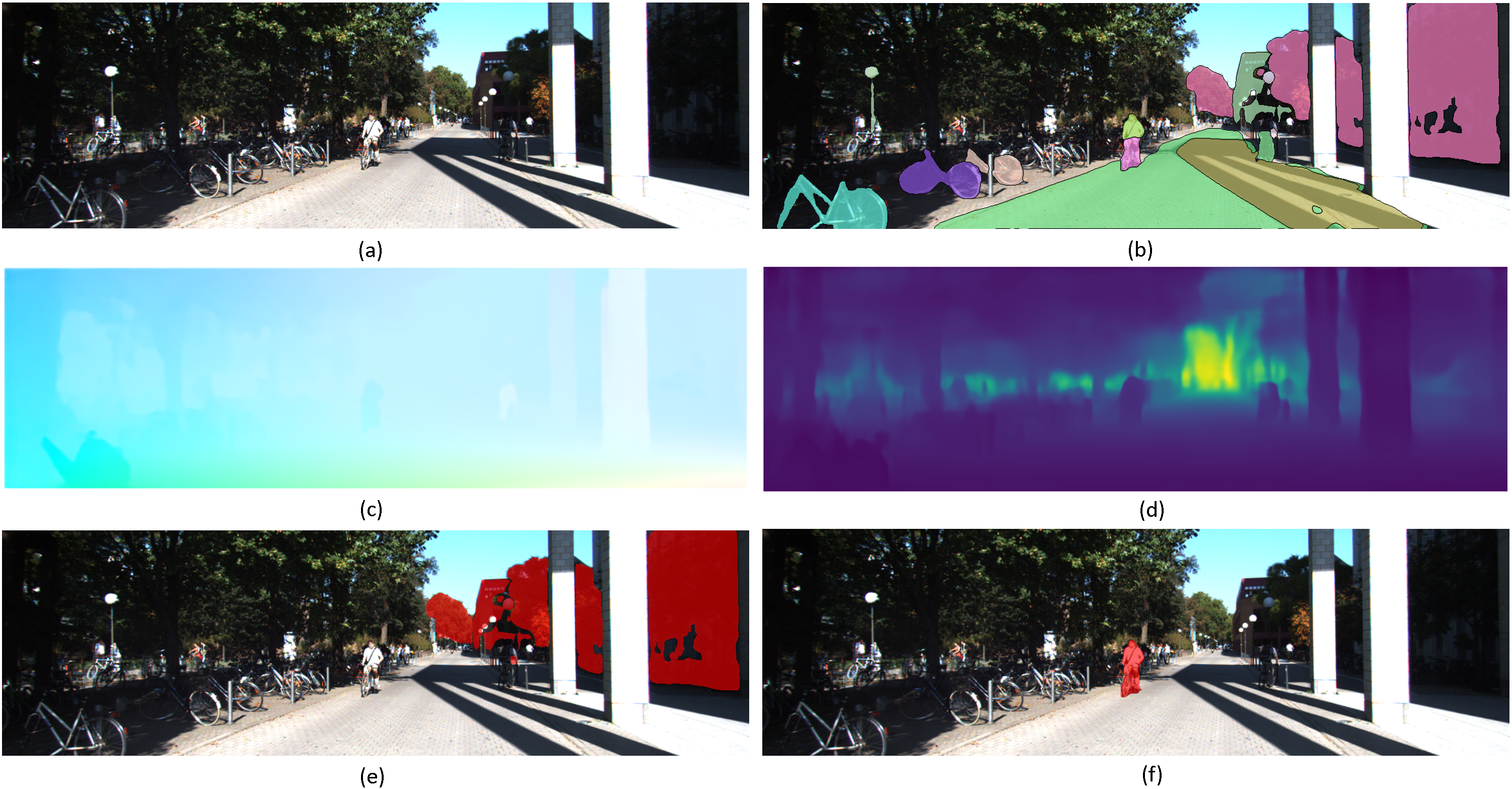}
    \end{subfigure}
    \caption{Motion segmentation results of the proposed method by using only optical flow vs. using both optical flow and relative depth as the motion cue. (a) is a frame from a video sequence. (b) is the object proposal generated for this input frame. (c) and (d) are the optical flow mask and the relative monocular depth map generated by the off-the-shelf deep learning models. (e) and (f) show the motion segmentation results of our method by using only optical flow and optical flow + relative depth. In this case, optical flow alone is insufficient to segment the moving object due to motion parallax as well as forward motion.}
    \label{fig: Overview of Outputs}
    \vspace{-0.2cm}
\end{figure*}

To illustrate why the overreliance on optical flow limits the performance of motion segmentation methods, figure \ref{fig: Overview of Outputs} shows an example of why optical flow by itself is insufficient if used as the only motion cue in motion segmentation. In this image sequence, the camera is moving forward and the cyclist is moving towards the camera. This particular image sequence has significant motion parallax since most objects in the scene are at different depths, which makes it almost impossible to tell which part of the scene is moving simply by looking at the optical flow field. This is because optical flow vectors are 2D projections of 3D velocity vectors and such projection is determined by both the depth and the screw motion of the object \cite{mitiche_computer_2014}. Therefore, when the optical flow is used as the only motion cue, the proposed method is not able to correctly segment the cyclist who is the only moving part in the scene. However, when the monocular depth map is jointly used with the optical field as motion cues, the method is able to produce the correct motion segmentation results.

To overcome these limitations of the existing methods, we propose a zero-shot monocular motion segmentation pipeline that is able to generate high-quality motion segmentation results without any training. We leverage the strong zero-shot capability of the computer vision foundation models to automatically recognize, detect, segment and track potential objects in the scene. Using this method, we can generate a high-quality object proposal for every frame in the image sequence. We then compute object-specific optical flow and depth map for each object on a per-frame basis. The depth map is generated by an off-the-shelf monocular estimation network, so it is only a pseudo depth map indicating relative depth. By analyzing how well each object's optical flow mask fits on every other objects given their relative depth, we are able to derive a motion affinity matrix representing the pairwise motion similarities between all pairs of objects in the object proposal. We then apply spectral clustering on the motion affinity matrix to cluster different objects into different motion groups and obtain the final motion segmentation.

Our method was evaluated on two widely recognized motion segmentation benchmark datasets: DAVIS-Moving and YTVOS-Moving. Although our method does not require any training, experiments show that is closely matched with the state-of-the-art supervised methods and is superior to the best unsupervised method. In summary, the key contributions are as follows: 
\begin{enumerate}
    \item We build the first zero-shot motion segmentation pipeline to achieve high-quality dense monocular motion segmentation without requiring any training.
    \item We show the effectiveness of using relative depth cue as a complement motion cue in improving optical flow based motion segmentation algorithms.
\end{enumerate}

%% file: 2_related_work.tex
\section{Related Work}

Dense monocular motion segmentation methods can generally be divided into two groups: (1) Intensity based methods \cite{negahdaripour_direct_1987, sekkati_variational_2007, cremers_detection_2009, papazoglou_v_2013, leibe_its_2016, bideau_best_2018} and (2) deep learning based methods \cite{vertens_smsnet_2017, siam_modnet_2018, bosch_deep_2021, ramzy_rst-modnet_2019, dave_towards_2019, leal-taixe_moa-net_2019, faisal_epo-net_2020, mohamed_monocular_2021, neoral_monocular_nodate, meunier_em-driven_2023}. Besides these two groups of methods which aim to generate a dense segmentation mask, there are also other motion segmentation methods that perform motion segmentation by clustering pre-computed point trajectories into different motion groups \cite{delong_fast_2010, isack_energy-based_2012, hutchison_object_2010, brox_object_2010, elhamifar_sparse_2013, ochs_segmentation_2014, lai_motion_2017, xu_motion_2018, vedaldi_usage_2020}. These methods do not belong to dense motion segmentation, but since they are relevant to our proposed method, we will briefly introduce them too. Additionally, it is important to differentiate between motion segmentation and video object segmentation (VOS) \cite{yao_video_2020}. The goal of VOS is to segment only the moving objects in the foreground, while motion segmentation focuses on segmenting any object that is moving independently, regardless of whether it is in the foreground or not.

\subsection{Intensity Based Methods}
Intensity based methods are traditional methods based on the image brightness constancy constraint, which presumes the pixel intensity of an image stays constant over a short period of time. Intensity based methods can be further categorized into indirect and direct methods. Indirect methods \cite{sekkati_variational_2007, cremers_detection_2009, papazoglou_v_2013, leibe_its_2016} use pixel-wise correspondences as input and generate a pixel-wise segmentation mask that delineates different motion groups. Such pixel correspondences are usually obtained from optical flow, which is based on the brightness constancy constraint. In contrast, direct methods \cite{aloimonos_direct_1984, negahdaripour_direct_1987, horn_direct_1988, vidal_closed_2005} directly take a pair of images as input and combine the two processes of optimizing for the brightness constancy constraint and estimating the motion models together. Most recent works on intensity based methods use optical flow based indirect methods, possibly due to the fast advance in optical flow estimation \cite{vedaldi_raft_2020, avidan_disentangling_2022}. Intensity based methods typically use causal inference or iterative optimization techniques to estimate the motion regions and motion models simultaneously \cite{sekkati_variational_2007, cremers_detection_2009, papazoglou_v_2013, leibe_its_2016}.

Intensity-based methods work well on simple scenes where the object motion and scene structure are simple, but will fail on more complex objects or scenes. For example, the motion of a walking human may not be modeled by a single optical flow mask since it may contain multiple moving parts (legs, arms and torso). In this case, these methods can have problems like over-segmentation or under-segmentation, where they segment different human parts as different motions, or only segment part of the human as the moving object. In addition, if the scene exhibits significant depth variation, these methods often struggle to determine whether a part of the image is moving independently or is simply located at a different depth compared to its surroundings.

\subsection{Deep Learning Based Methods} 
Most deep learning methods take a sequence of image frames and sometimes also the precomputed motion cues such as the optical flow field or the monocular depth maps of these images, and produce a dense segmentation map in an end-to-end manner. Earlier methods are often only able to perform binary motion segmentation \cite{vertens_smsnet_2017, siam_modnet_2018, ramzy_rst-modnet_2019, faisal_epo-net_2020, bosch_deep_2021}, but more recent methods have made significant progress and are able to achieve promising results in multi-label motion segmentation \cite{dave_towards_2019, mohamed_monocular_2021, neoral_monocular_nodate, yang_learning_2021, choudhury_guess_2022, meunier_em-driven_2023}.

Many deep learning based methods adopt a fully supervised approach \cite{vertens_smsnet_2017, siam_modnet_2018, dave_towards_2019, mohamed_monocular_2021}. These methods typically train a CNN-based encoder-decoder network to perform end-to-end learning, which is computation-intensive. Their network architecture usually have the following components: (1) a module to extract the motion information from consecutive image frames, (2) a module to extract appearance information from the same sequence of frames, (3) a module to fuse the appearance and motion information together, and (4) a decoder to generate the final segmentation. These methods perform very well on scenes similar to the datasets they are trained on, but cannot scale well to unseen environment where there are different motion patterns or object classes. Moreover, the data collection and training process are very time-consuming and computation intensive, which make them not an ideal method. 

Aside from supervised methods, some methods also use semi-supervised, self-supervised and unsupervised approaches. In \cite{leal-taixe_moa-net_2019}, the authors extended their previous work \cite{leibe_its_2016} by proposing an self-supervised approach to train a neural network to perform motion segmentation on synthetic angle fields, given that most optical flows can be reduced to rotation-compensated angle fields. In \cite{meunier_em-driven_2022}, the authors proposed an unsupervised learning method to solve multi-label motion segmentation problem by training a neural network to mimic the motion segmentation results from the Expectation-Maximization (EM) algorithm. However, these two methods purely rely on optical flow for motion information and thereby inheriting its limitations. To alleviate this problem, \cite{choudhury_guess_2022} proposes to train image segmentation and motion segmentation models together using both optical flow and raw video frames as inputs due to the fact that motion and appearance cues are usually highly related in practice. The unsupervised training is done in a very similar way as \cite{meunier_em-driven_2022} using the EM-algorithm.

\subsection{Sparse Correspondence Based Methods}
Unlike Intensity based methods or deep learning methods, sparse correspondence based methods cannot produce dense segmentation masks of different moving objects. Instead, they output clusters of predefined kepypoints corresponding to different motion groups instead of dense segmentation masks. These methods can be further categorized into two-frame based methods and multi-frame based methods. Two-frame based methods \cite{lasenby_geometric_1998, delong_fast_2010, isack_energy-based_2012, barath_progressive-x_2019} typically determine motion parameters using iterative optimization methods. This involves identifying a specific number of geometric models, such as homography, based on a set of corresponding keypoints, aiming to minimize an energy function representing the overall quality of the corresponding keypoints clustering. Unlike two-frame based methods, multi-frame based methods \cite{lai_motion_2017, xu_motion_2018, vedaldi_usage_2020, jiang_what_2021, xi_multi-motion_2022, elhamifar_sparse_2013, rao_motion_2010, tron_benchmark_2007, vidal_subspace_2011, brox_object_2010, ochs_segmentation_2014} usually establish point correspondences over multiple frames using an optical flow based point tracker. Noisy, occluded and unwanted points are often manually removed to produce a sparse set of completely noise-free point trajectories. Multi-frame based methods have proven to be superior than two-frame based methods due to their ability to analyze the motion data in a longer time window using various geometric models and spatio-temporal similarities. Moreover, unlike two-frame based methods which only rely on epipolar geometry to detect different motions, multi-frame based methods are able to combine different geometric models together and achieve impressive results in challenging scenes with complex motions.

%% file: 4_methodology.tex
\section{Motion Segmentation Pipeline}

\begin{figure*}[t]  
    \vspace{-0.2cm}
    \centering
    \includegraphics[width=\linewidth]{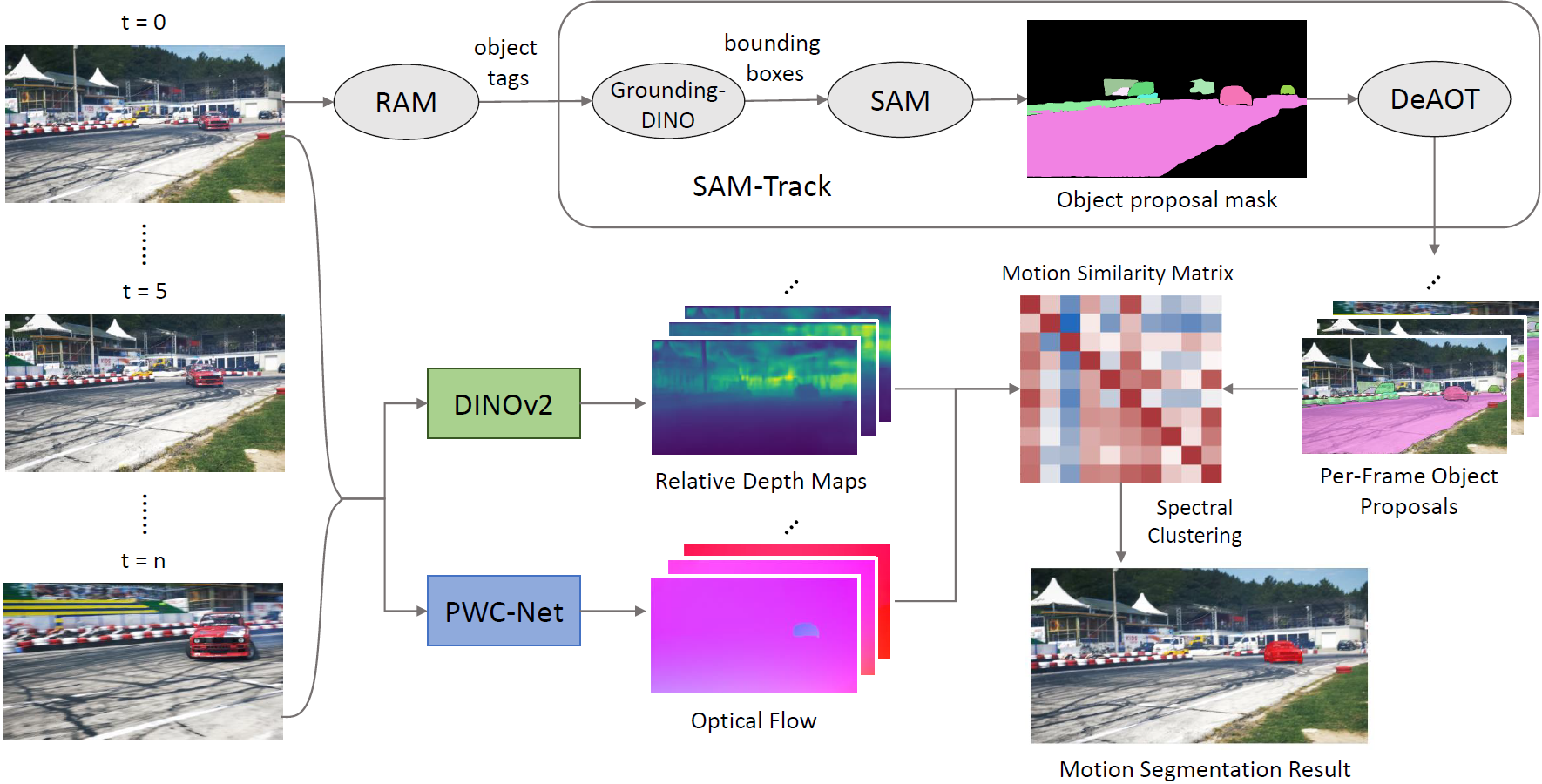} 
    \caption{Diagram of our proposed motion segmentation method. Given an image sequence, the proposal extraction module automatically identifies, segments and tracks all common objects through the whole sequence to generate an object proposal for each frame. Meanwhile, the motion cue generation module generates optical flow masks and monocular depth maps using PWC-Net and DINOv2. object-specific optical flow and depth maps are obtained by combining object proposals with optical flow and monocular depth maps. Given these object-specific motion cues, pairwise object similarity scores are computed to consturct the motion similarity matrix. Finally, spectral clustering is used to cluster each object into its motion group.}
    \label{fig: Motion Segmentation Pipeline}
    \vspace{-0.2cm}
\end{figure*}

We introduce a novel monocular dense motion segmentation pipeline that performs . The method first automatically extracts high-quality object proposals using the computer vision foundation models, and then clusters these proposed objects into distinct motion groups according the motion cues provided by the object-specific optical flow and relative depth maps. Figure \ref{fig: Motion Segmentation Pipeline} shows a diagram of our motion segmentation pipeline.

\subsection{Automatic Object Proposal Extraction}
To automatically identify all moving objects within a video, the first task is to recognize and detect each common object in the video and track their trajectories throughout the video. We leverage the recently proposed computer vision foundational models in object recognition (RAM)\cite{zhang_recognize_2023}, detection (Grounding DINO)\cite{liu_grounding_2023}, and segmentation (SAM)\cite{rajic_segment_2023}, alongside a state-of-the-art object tracking model (DeAOT) \cite{yang_decoupling_2022}. This preprocessing pipeline is based on the Segment and Track Anything (SAMTrack) \cite{cheng_segment_2023} framework, which integrates the aforementioned models into a unified object segmentation and tracking system. SAMTrack generates dense object tracking masks based on a user-defined textual prompt specifying the desired objects to be tracked. To automate our system and eliminate the need for manual text prompts, we incorporate RAM at the beginning of our pipeline to automatically identify common objects in the initial video frame. 

In summary, our complete preprocessing pipeline involves using RAM to identify common objects in the first frame, using the output from RAM as a textual prompt for the Grounding DINO model to obtain object bounding boxes, and then using these bounding boxes with SAM to generate an instance segmentation mask of the first frame. Non-max suppression is applied to remove overlapping objects with an IoU score greater than 0.5 or with a mask area exceeding half the image size. Finally, The DeAOT tracker is used to follow each object’s mask throughout the video. 

To account for potential new objects entering the scene midway through the video, we divide the video into multiple parts consisting of equal numbers of frames and apply the full preprocessing pipeline to each part individually. The specific length of the each individual part can vary, but it typically more beneficial to set smaller number in more dynamic videos where more objects enter the scene midway.

\subsection{Optical Flow and Relative Depth as Object-Specific Motion Cues}
In order to determine if a set of objects have the same motion, we need to analyze object-specific motion cues for each object. More specifically, we calculate a dense optical flow mask and a monocular relative depth map for every video frame as the motion cues. The extraction of optical flow masks and relative depth maps are conducted using PWC-Net \cite{sun_disentangling_2022} and DINOv2 \cite{oquab_dinov2_2023} respectively. Both models are the state-of-the-art models in their respective domains. 

Relying solely on optical flow for motion segmentation is inadequate, as it cannot effectively distinguish between different motions and different depths. This limitation becomes evident when a camera moves, causing two stationary objects at different depths to appear as if they are moving differently due to motion parallax. To address this limitation, it is essential to integrate depth information with optical flow, enhancing its ability to accurately analyze motion. In the following section, we present a parametric model that combines optical flow and depth data. We will demonstrate how this model can be used to compute the three-dimensional screw motions of objects, thus enabling the differentiation of objects based on their motions. This parametric model offers a robust theoretical framework for understanding complex motions in dynamic scenes.

\subsection{Parametric Motion Model}
We propose a motion model fitting algorithm that uses a parametric model derived from optical flow and depth data to represent the motions of individual objects throughout the video. This parametric motion model incorporates a revised version of the model equations introduced by Longuet-Higgins and Pruzdny \cite{longuet-higgins_interpretation_1980}, which can be used to compute the instantaneous screw motion of rigid objects at arbitrary depths. The original Longuet-Higgins and Pruzdny model equations establish a relationship between the optical flow, the instantaneous screw motion of rigid objects and the depths of individual pixels as follows: 

\vspace{-0.01cm}
\begin{equation} \label{eq:1}
\begin{split}
u = -\frac{xy}{f}\omega_1 + \frac{f^2 + x^2}{f}\omega_2 - y\omega_3 + \frac{f\tau_1 - x\tau_3}{z} 
\\[1ex]
v = -\frac{f^2 + y^2}{f}\omega_1 + \frac{xy}{f}\omega_2 + x\omega_3 + \frac{f\tau_2 - y\tau_3}{z}
\end{split}
\end{equation}
\vspace{0.01cm}

Here, $u$ and $v$ denote the optical flow vectors along the x and y axes respectively, $z$ denotes pixel depth, $f$ stands for the camera's focal length, and $\tau_1$, $\tau_2$, $\tau_3$, $\omega_1$, $\omega_2$, $\omega_3$ symbolize the object's translational and rotational movements. Nonetheless, the absolute depth of each pixel is often unknown in practice, making the complete utilization of this model for computing object motion unfeasible. To overcome this limitation, existing methods often use a parametric equation to infer object motion directly from the optical flow without knowing depth. For instance, \cite{meunier_em-driven_2023} applies a segmented parametric formula with 12 parameters to precisely align with the optical flow field:

\begin{equation} \label{eq:2}
\begin{split}
u = a + b x + c y + d x^2 + e xy + f y^2
\\[1ex]
v = g + h x + i y + j x^2 + k xy + l y^2
\end{split}
\end{equation}
\vspace{0.01cm}

However, such a motion model lacks theoretical accuracy and fails to accommodate scenes with significant depth variations. Other research \cite{leibe_its_2016, bideau_best_2018} employs a simpler parametric motion equation to estimate the rotation-compensated optical flow angle field, albeit requiring known camera intrinsic parameters, which is impractical. To establish a motion model that is both theoretically robust and independent of camera intrinsics, we propose to linearize the Longuet-Higgins and Pruzdny equations using the monocular depth map produced from DINOv2. With the relative depth of each pixel known, we can reformulate the original equations into the following linear parametric form:

\begin{equation} \label{eq:3}
\begin{split}
u = a + b \frac{1}{z} - c \frac{x}{z} -dy + ex^2 - fxy
\\[1ex]
v = g + h \frac{1}{z} - c \frac{y}{z} -dx + exy + fy^2
\end{split}
\end{equation}
\vspace{0.01cm}

This set of linearized equations is more robust to motion parallax than (2) where the depth value is encoded in multiple unknown parameters and need to be approximated together with the screw motions. Using relative depth is sufficient for our goal, which is to distinguish different motions instead of computing the absolute values of the screw motion parameters.

\subsection{Pairwise Motion Similarity Matrix}
Once all optical flow motion models are obtained, each object will have a parametric motion model for every pair of frames. By fitting every object’s optical flow vectors and depth map on its parametric motion model for the same frame pair, we can compute the residuals of each object to the motion models of all other objects. The residuals are computed with the mean squared error. Given {\it N} objects proposed for the scene in total, the motion residual vector for the {\it i}-th object at frame pair {\it m} are derived as follows:
\begin{equation*} \label{eq:4}
\setlength{\jot}{10pt} 
\begin{split}
{\pmb e}_i^m = [{e}_{i,1}^m, {e}_{i,2}^m, {e}_{i,3}^m, ..., {e}_{i,N}^m], 
\end{split}
\end{equation*}
\vspace{0.1cm}
where ${e}_{i,n}^m$ represents the mean residual calculated by fitting the motion model of object $i$ to the optical flow and depth of object $n$ for frames $m$ and $m + 1$. A motion similarity matrix is then constructed to encode the motion similarity scores across all pairs of objects. This is achieved using the ordered residual kernel (ORK) \cite{chin_ordered_2009}. To do so, the residual values in each residual vector are first sorted in ascending order. A threshold {\it t} is set to select the lowest {\it t}-th residuals as inliers. An binary inlier vector ${\pmb v_{i}} = \{0,  1\}^N$ is then obtained for each object, where $N$ is the total number of objects. The pairwise motion similarity between objects $i$ and $j$ can be calculated as ${\pmb d_{ij} = \pmb v_{i}^\intercal \pmb v_{j}}$, representing the frequency at which these two objects occur as each other's motion inliers. The ORK selects a certain number of inliers from each object's motion vectors instead selecting inliers below a threshold, making it more robust to different scenes and motions. After the motion similarity matrix is constructed, each motion similarity score ${\pmb d_{ij}}$ is normalized by dividing it with the number of frames that objects $i$ and $j$ have in common. This normalization step helps eliminate the weighting bias caused by incomplete trajectories.

\subsection{Clustering Objects into Distinct Motion Groups}
We apply row normalization \cite{von_luxburg_tutorial_2007} to the constructed motion similarity matrix and use spectral clustering to cluster objects into different motion groups. Given a predefined number of motion groups, spectral clustering cluster objects with high motion similarity values into the same motion group. Spectral clustering is a widely adopted technique in sparse correspondence based motion segmentation methods. It has shown effectiveness in clustering different point trajectories into different motion groups \cite{xu_motion_2018, huang_motion_2023, jiang_what_2021}. Therefore, we use it to cluster different objects in the object proposal into different motion groups given the motion similarity matrix. 

%% file: 5_experiments.tex
\newcolumntype{Y}{>{\centering\arraybackslash}X}
\newcolumntype{C}[1]{>{\centering\arraybackslash}m{#1}} 
\newcolumntype{L}[1]{>{\raggedright\arraybackslash}m{#1}} 
\newcolumntype{B}[1]{>{\raggedright\arraybackslash}b{#1}}
\newenvironment{Table}[4]{%
    \longtable{%
        |>{\centering}A{#4}{1.5}
        |}\hline\ignorespaces}{%
\endlongtable\ignorespacesafterend}

\begin{figure*} [thbp!]  
    \vspace{-0.2cm}
    \centering
    \includegraphics[width=\textwidth]{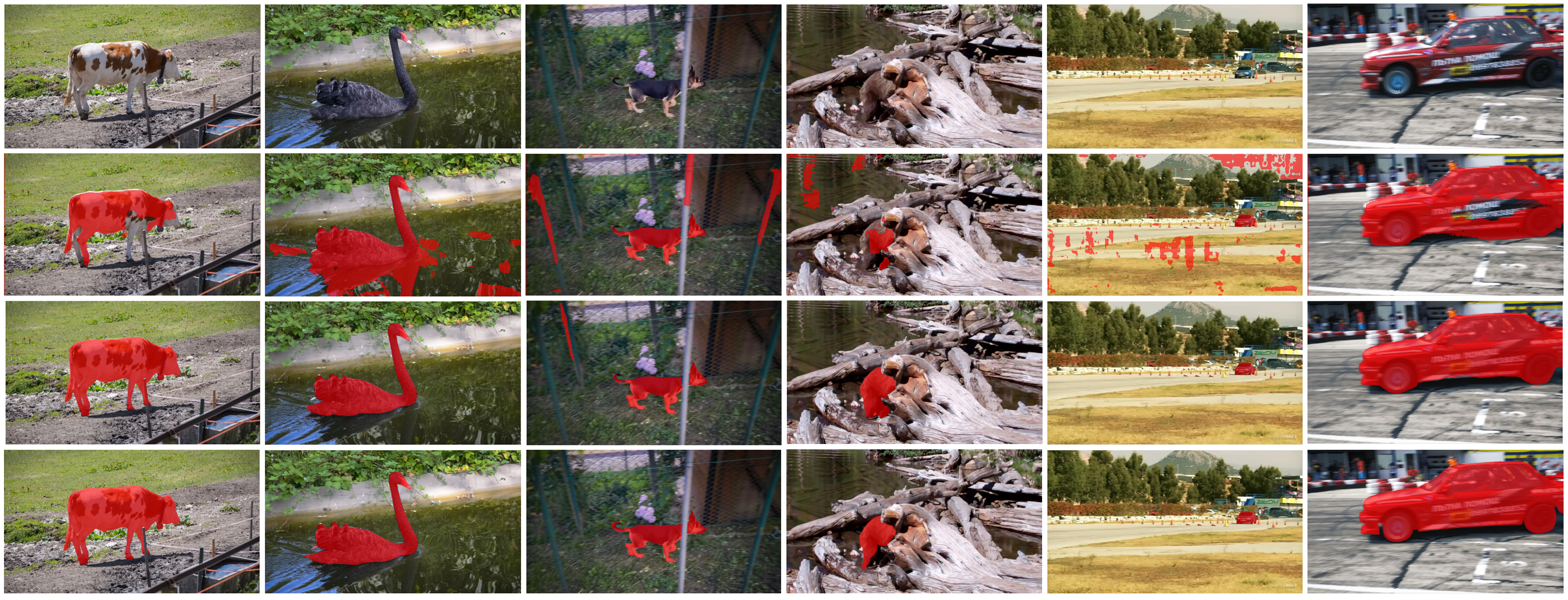} 
    \caption{Qualitative comparison with the state-of-the-art unsupervised dense motion segmentation method \cite{meunier_em-driven_2023} on DAVIS-Moving (column 1 - 3) and YTVOS-Moving (column 4 - 6) datasets. First row: Original video frame. Second row: Motion segmentation results produced by \cite{meunier_em-driven_2023}. Third row: Motion segmentation results of our method. Last row: Ground truth}
    \label{fig: Binary MoSeg Results}
\end{figure*}

\section{Experiments}

We conduct experiments using two widely recognized datasets: DAVIS-Moving and YTVOS-Moving. This section provides an overview of these datasets and the evaluation metrics, and compares our method with state-of-the-art approaches. Additionally, we present an ablation study to analyze the contributions of optical flow and depth information in enhancing motion segmentation results compared to the baseline. 

\subsection{Datasets and Evaluation Metrics}
DAVIS-Moving and YTVOS-Moving \cite{dave_towards_2019} are the most recent benchmarks for generic instance motion segmentation. They are subsets of the DAVIS-17 dataset \cite{pont-tuset_2017_2018} and the YTVOS dataset \cite{ferrari_youtube-vos_2018} respectively. On the contrary to the original DAVIS and YTVOS datasets which focus on video object segmentation and only label moving objects in the foreground, the DAVIS-Moving and YTVOS-Moving datasets contain only videos where all moving objects are labeled. These datasets are particularly challenging due to occlusions, non-rigid motions and the diversity of object classes. We evaluate our method using the precision (Pu), recall (Ru), and F-score (Fu) metrics, as proposed by \cite{dave_towards_2019}. These metrics are designed to penalize false positives, with the F-score providing an overall performance score by combining precision and recall.

\subsection{Results}

Tables \ref{table: Binary MoSeg DAVIS-Moving} and \ref{table: Binary MoSeg YTVOS-Moving} present quantitative comparisons of our method with the state-of-the-art unsupervised motion segmentation method (EM) \cite{meunier_em-driven_2023} on DAVIS-Moving and YTVOS-Moving datasets. The comparisons are limited to binary motion segmentation due to the availability of model weights from \cite{meunier_em-driven_2023}. However, these results are indicative of performance on multi-label segmentation tasks. Our method demonstrates superior performance on both datasets, producing higher F-scores than EM, showing more accurate segmentation results. 

Figure \ref{fig: Binary MoSeg Results} shows a qualitative comparison between our method and EM on the DAVIS-Moving and YTVOS-Moving datasets. The first row shows the original video frames, the second row shows the results from EM, the third row shows the results from out method, and the last row shows the ground truth. Both methods can detect moving objects in the scene, but our method produces much more coherent object masks and produces significantly less false-positive segments.

\begin{table}[h!t!b!p!]
    \renewcommand{\arraystretch}{1.6} 
    \centering
    \begin{tabularx}{\columnwidth}{| L{2cm} | L{2.5cm} | YYY |}
        \hline
        {\raggedright \textbf{Method}} & 
        {\raggedright \textbf{Training Method}} & 
        \textbf{Pu} & \textbf{Ru} & \textbf{Fu} \\
        \hline
        EM \cite{meunier_em-driven_2023} & Unsupervised & 61.29 & \textbf{86.56} & 68.90  \\
        \hline
        \raggedright \textbf{Ours} & 
        \raggedright \makecell[{{p{2cm}}}]{ \vspace{0.1pt} Zero-Shot \\ (no training) } \vspace{0.1cm} &
        \textbf{74.47} & 
        77.78 & 
        \textbf{75.96} \\
        \hline
    \end{tabularx}
    \caption{Quantitative binary motion segmentation results of our method and the state-of-the-art unsupervised motion segmentation method (EM) \cite{meunier_em-driven_2023} on DAVIS-Moving. Our method significantly outperforms EM with a much higher Fu score.}
    \label{table: Binary MoSeg DAVIS-Moving}
\end{table}

\begin{table}[h!t!b!p!]
    \renewcommand{\arraystretch}{1.6} 
    \centering
    \begin{tabularx}{\columnwidth}{|L{2cm} | L{2.5cm} | YYY |}
        \hline
        {\raggedright \textbf{Method}} & 
        {\raggedright \textbf{Training Method}} & 
        \textbf{Pu} & \textbf{Ru} & \textbf{Fu} \\
        \hline
        EM \cite{meunier_em-driven_2023} & Unsupervised & 41.78 & 39.06 & 35.38  \\
        \hline
        \raggedright \textbf{Ours} & 
        \raggedright \makecell[{{p{2cm}}}]{\vspace{0.1pt} Zero-Shot \\ (no training)} \vspace{0.1cm} & 
        \textbf{54.63}  & 
        \textbf{50.36}  & 
        \textbf{50.78}  \\
        \hline
    \end{tabularx}
    \caption{Quantitative binary motion segmentation results of our method and EM \cite{meunier_em-driven_2023} on the YTVOS-Moving dataset. Our method still significantly outperforms EM.}
    \label{table: Binary MoSeg YTVOS-Moving}
\end{table}

\begin{table*}[h!t!b!p!]
    \vspace{-0.1cm}
    \renewcommand{\arraystretch}{1.6} 
    \centering
    \begin{tabularx}{\textwidth}{|L{2cm} | L{3cm} |YYY |YYY |}
        \hline
        \multirow{2}{*}{\raggedright \makecell{\textbf{Method}}} & 
        \multirow{2}{*}{\centering \makecell{\textbf{Training Method}}} & 
        \multicolumn{3}{c|}{\textbf{DAVIS-Moving}} & 
        \multicolumn{3}{c|}{\textbf{YTVOS-Moving}} \\ 
        \textbf{} & \textbf{} &
        \textbf{Pu} & \textbf{Ru} & \textbf{Fu} & 
        \textbf{Pu} & \textbf{Ru} & \textbf{Fu} \\
        \hline
        MoSeg \cite{dave_towards_2019} & Supervised & 78.30 & 78.80 & 78.10 & 74.50 & 66.40 & 66.38 \\
        \hline
        Raptor \cite{neoral_monocular_nodate} & \multirow{2}{*}{\raggedright \makecell{Supervised Features}} & 75.90 & 79.67 & 75.93 & 64.43 & 60.94 & 60.35 \\
        RigidMask \cite{yang_learning_2021} & \textbf{} & 59.03 & 49.89 & 50.01 & 29.88 & 17.88 & 18.70 \\
        \hline
        Ours & \textbf{Zero-Shot (no training)} & 71.53 & 75.66 & 73.18 & 63.54 & 58.94 & 56.06 \\
        \hline
    \end{tabularx}
    \caption{Quantitative results comparing our method with state-of-the-art supervised and semi-supervised methods on the DAVIS-Moving and YTVOS-Moving validation datasets show that while our method lags behind the supervised MoSeg, it matches semi-supervised methods and even outperforms RigidMask, despite no training.}
    \label{table: Quantitative Results and Ablation Study}
\end{table*}

Table \ref{table: Quantitative Results and Ablation Study} shows quantitative comparison of our method and state-of-the-art supervised and semi-supervised methods on the two benchmarks. Both binary and multi-label motion segmentation scenes are used. Although we did not conduct any training, our method achieves competitive results and even outperforms one of the semi-supervised methods (RigidMask).

\begin{figure*} [thbp!]  
    \centering
    \includegraphics[width=\textwidth]{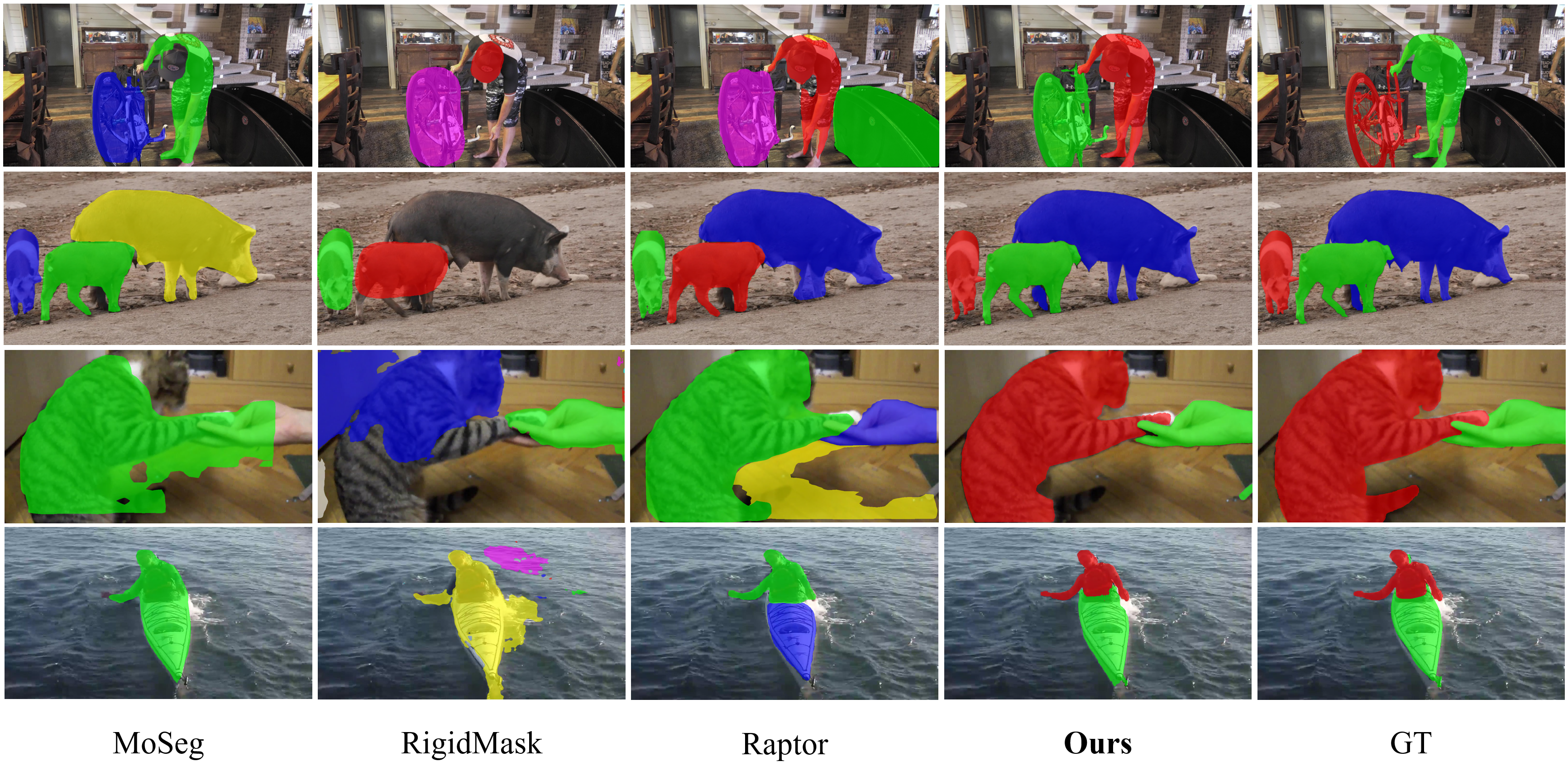} 
    \caption{Qualitative comparison with state-of-the-art methods on DAVIS-Moving (rows 1-2) and YTVOS-Moving (rows 3-4). MoSeg performs best as a supervised method. RigidMask struggles with non-rigid motions, and Raptor has similar issues but to a lesser extent. Our method matches the performance of the supervised method in these challenging scenarios.}
    \label{fig: multilabel MoSeg Results}
\end{figure*}

We also show qualitative comparison with these methods in Figure \ref{fig: multilabel MoSeg Results}. MoSeg\cite{dave_towards_2019} produces the best results on both datasets. RigidMask\cite{yang_learning_2021} fails to produce coherent segmentation on most scenes, and also fails to detect the motions of the parrot and the train on the last two rows, whereas our method successfully detects and segments them coherently. Raptor\cite{neoral_monocular_nodate} is able to detect most objects in the scene thanks to its powerful semantic backbone, but it still over-segments non-rigid objects like the parrot. Our method performs well in these cases, performing almost as well as the supervised method. 

\begin{table*}[h!t!b!p!]
    \renewcommand{\arraystretch}{1.6} 
    \centering
    \begin{tabularx}{\textwidth}{|L{3cm} | YYY |YYY |}
        \hline
        \multirow{2}{*}{\raggedright \makecell{\textbf{Method}}} & 
        \multicolumn{3}{c|}{\textbf{DAVIS-Moving}} & 
        \multicolumn{3}{c|}{\textbf{YTVOS-Moving}} \\
        \textbf{} &
        \textbf{Pu} & \textbf{Ru} & \textbf{Fu} & 
        \textbf{Pu} & \textbf{Ru} & \textbf{Fu} \\
        \hline
        OC + Depth & \textbf{71.53} & 75.66 & \textbf{73.18} & \textbf{63.54} & 58.94 & \textbf{56.06} \\
        \hline
        OC  & 58.25 & 59.22 & 57.08 & 61.79 & 54.64 & 53.74  \\
        \hline
        Base (obj. proposal) & 43.17 & \textbf{86.24} & 52.12 & 48.49 & \textbf{73.01} & 50.82 \\
        \hline
    \end{tabularx}
    \caption{Quantitative ablation study: Motion segmentation results from only using optical flow (OC) as motion cue vs. using both optical flow and depth map (OC + Depth) as motion cue. Baseline results (Base) are also shown. Baseline is obtained by directly using the raw object proposals as the final motion segmentation mask. Bold numbers are the best results.}
    \label{table: Quantitative Ablation Study}
\end{table*}

\begin{figure*} [thbp]  
    \centering
    \includegraphics[width=\textwidth]{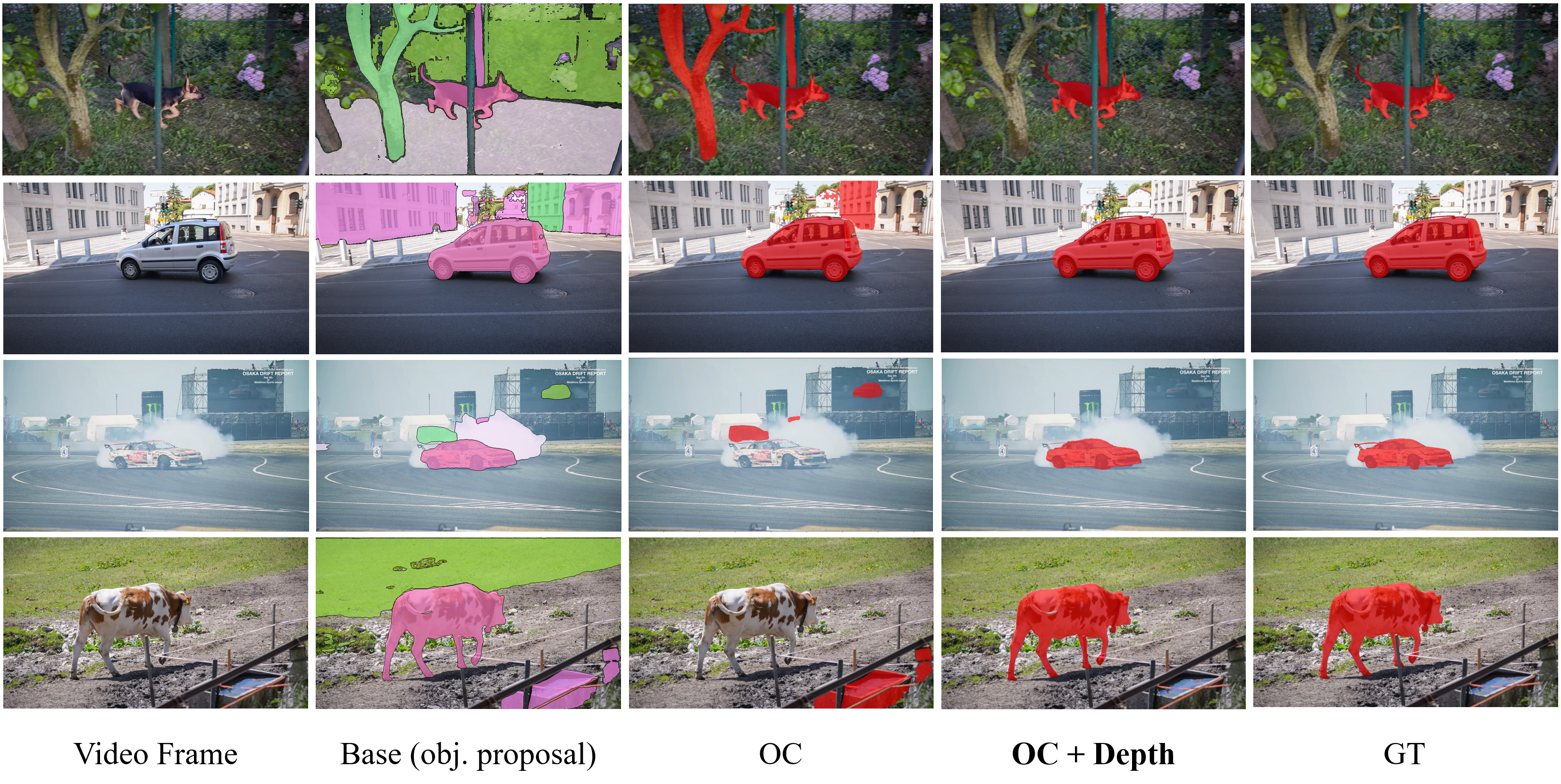} 
    \caption{Qualitative ablation study: Qualitative comparison between motion segmentation results using optical flow alone (OC) and both optical flow and depth map (OC + Depth). Pure optical flow based motion model (OC) suffers when multiple objects are at different depths. Combining optical flow with depth (OC + Depth) significantly mitigates this problem.}
    \label{fig: Ablation Study}
\end{figure*}

Our method has two main limitations: slow inference speed due to multiple deep learning models, and the need for a predefined number of motions in the scene. The latter can be addressed by incorporating an automatic model selection technique \cite{huang_unified_2023}. The inference speed, however, can only be improved by training an end-to-end network using our proposed motion residual functions.

\subsection{Ablation Study}

We present an ablation study to compare the performances of the two different motion models with the baseline, which is obtained from the raw object proposal alone. Both qualitative (Figure. \ref{fig: Ablation Study}) and quantitative (Table \ref{table: Quantitative Ablation Study}) comparisons are shown between our motion segmentation result obtained by only using optical flow as the motion cue vs. using both optical flow and the depth map, as well as the baseline results.

To produce the motion segmentation results only from optical flow, We use the optical flow motion model of EM \cite{meunier_em-driven_2023}, which is the state-of-the-art unsupervised motion segmentation method using only optical flow as input. Their 12-parameter quadratic parametric motion model integrates the unknown depth information as unknown parameters (as shown in equation \eqref{eq:2}). Results indicate that the motion model integrating both optical flow and depth (OC + Depth) significantly outperforms the model relying solely on optical flow (OC) across all metrics on DAVIS-Moving. However, on the YTVOS-Moving dataset, the performance improvement is small, suggesting that unknown depth information is not a critical limiting factor. This can be attributed to several reasons: First, many objects labeled as moving in YTVOS-Moving are static in most frames. Second, the dataset includes significant occlusions and uncommon objects such as camouflaged animals. These challenges likely have a greater impact on the accuracy of motion segmentation than the absence of depth information. 

%% file: 6_conclusions.tex
\section{Conclusion}
We introduce a novel approach for dense monocular motion segmentation that operates without requiring any training. By integrating the deep learning models with traditional optical flow-based methods, we propose a zero-shot technique that effectively clusters object proposals into distinct motion groups. Our method enhances the performance of conventional optical flow-based techniques by incorporating monocular depth maps, resulting in superior outcomes compared to using optical flow alone. Remarkably, despite the absence of training, our approach surpasses the state-of-the-art unsupervised motion segmentation methods on two widely-adopted benchmarks and rivals the top supervised and semi-supervised methods. 

Future work will focus on enhancing our method by integrating additional motion cues and geometric models, such as keypoint correspondences and the fundamental matrix, to further boost its performance, as well as incorporating a model selection method to automatically infer the number of motions in the scene.